\documentclass[journal]{IEEEtran}
\ifCLASSINFOpdf
\else
\fi

\usepackage{graphicx}
\usepackage{caption}
\usepackage{subcaption}
\usepackage{multirow} 
\usepackage{multicol}
\usepackage{color}
\usepackage{comment}

\begin{document}
%
\title{Enabling 3D Object Detection with a Low-Resolution LiDAR}
%
%
%

\author{Lin Bai,~\IEEEmembership{Student~Member,~IEEE,}
        Yiming~Zhao,~\IEEEmembership{Member,~IEEE,}
        and~Xinming~Huang,~\IEEEmembership{Senior~Member,~IEEE}
\thanks{*This work was supported in part by the US National Science Foundation Grant 2006738 and by The MathWorks.}
\thanks{Lin Bai, Yiming Zhao and Xinming Huang are with the Department of Electrical and Computer Engineering, Worcester Polytechnic Institute, Worcester, MA 01609, USA, email: \{lbai2, yzhao7, xhuang\}@wpi.edu}%
}

%
%

\markboth{Journal of \LaTeX\ Class Files,~Vol.~14, No.~8, August~2015}%
{Shell \MakeLowercase{\textit{et al.}}: Bare Demo of IEEEtran.cls for IEEE Journals}
%



\maketitle

\begin{abstract}
Light Detection And Ranging (LiDAR) has been widely used in autonomous vehicles for perception and localization. However, the cost of a high-resolution LiDAR is still prohibitively expensive, while its low-resolution counterpart is much more affordable. Therefore, using low-resolution LiDAR for autonomous driving is an economically viable solution, but the point cloud sparsity makes it extremely challenging. In this paper, we propose a two-stage neural network framework that enables 3D object detection using a low-resolution LiDAR. Taking input from a low-resolution LiDAR point cloud and a monocular camera image, a depth completion network is employed to produce dense point cloud that is subsequently processed by a voxel-based network for 3D object detection. Evaluated with KITTI dataset for 3D object detection in Bird-Eye View (BEV), the experimental result shows that the proposed approach performs significantly better than directly applying the 16-line LiDAR point cloud for object detection. For both easy and moderate cases, our 3D vehicle detection results are close to those using 64-line high-resolution LiDARs.
\end{abstract}

\begin{IEEEkeywords}
\textcolor{black}{Low-resolution LiDAR, Camera, 3D Vehicle Detection.}
\end{IEEEkeywords}

%
\IEEEpeerreviewmaketitle

\section{INTRODUCTION}
\begin{figure*}[htbp]
    \captionsetup[subfigure]{aboveskip=-0.5pt,belowskip=-1pt}
    \centering
    \begin{subfigure}{0.32\textwidth}
        \includegraphics[width=0.98\textwidth]{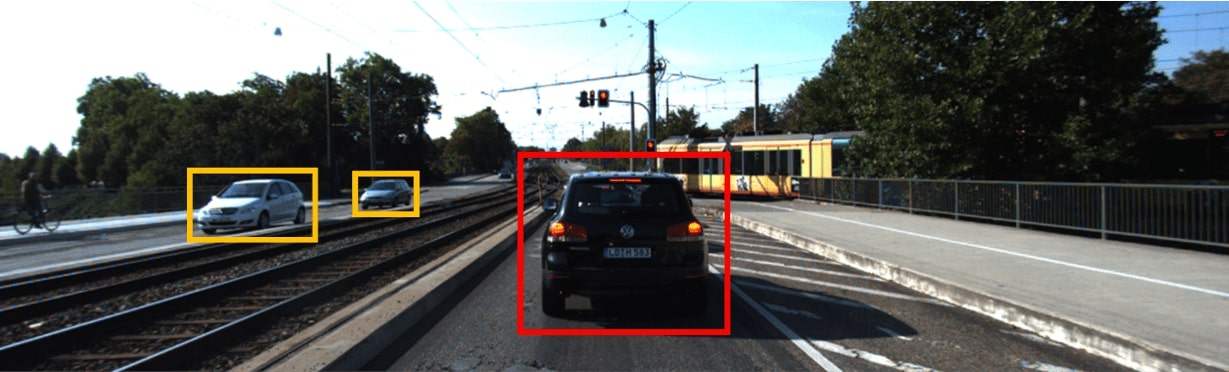}
        \caption{image scenario 1}
    \end{subfigure}%
    \begin{subfigure}{0.32\textwidth}
        \includegraphics[width=0.98\textwidth]{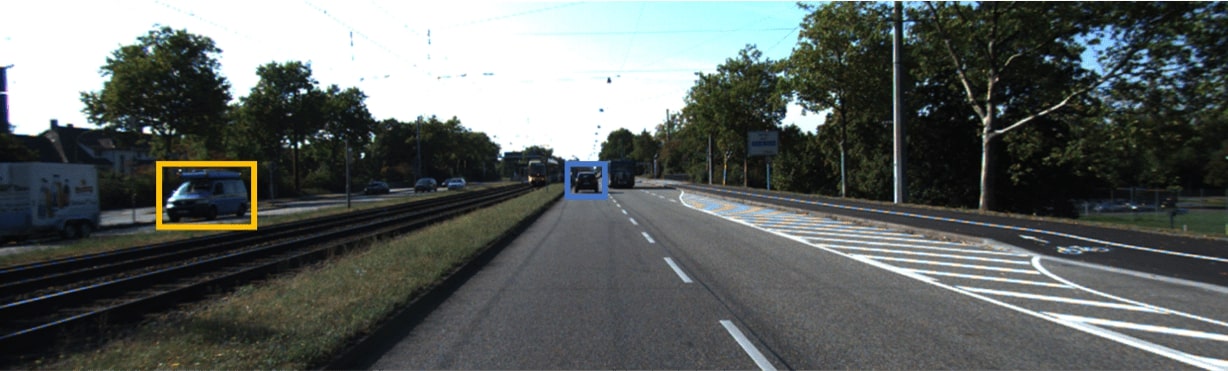}
        \caption{image scenario 2}
    \end{subfigure}%
    \begin{subfigure}{0.32\textwidth}
        \includegraphics[width=0.98\textwidth]{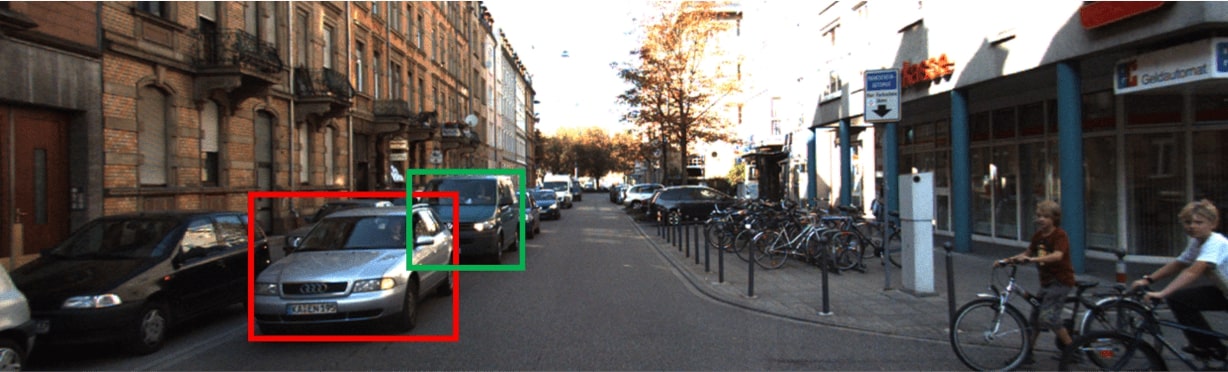}
        \caption{image scenario 3}
    \end{subfigure}%
    \newline
    \begin{subfigure}{0.32\textwidth}
        \includegraphics[width=0.98\textwidth]{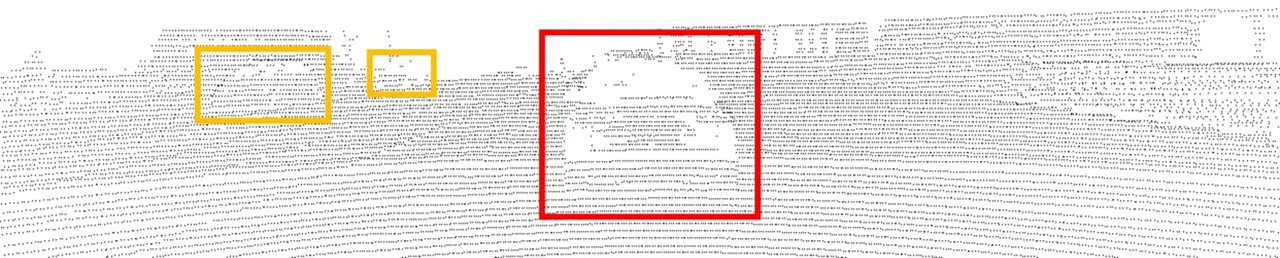}
        \caption{dense depth map scenario 1}
    \end{subfigure}%
    \begin{subfigure}{0.32\textwidth}
        \includegraphics[width=0.98\textwidth]{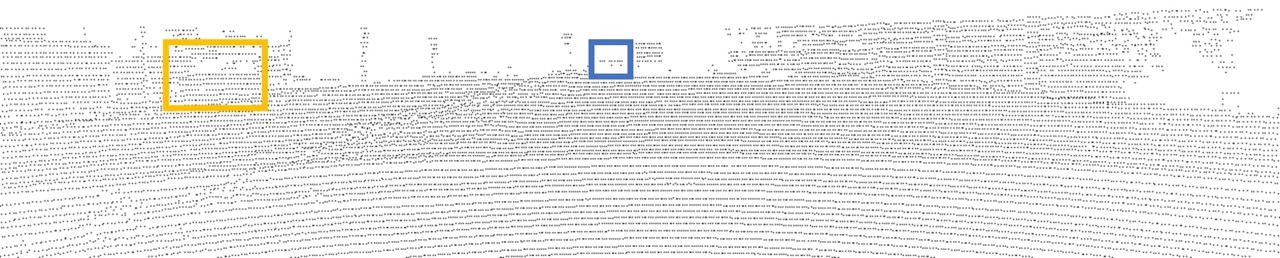}
        \caption{dense depth map scenario 2}
    \end{subfigure}%
    \begin{subfigure}{0.32\textwidth}
        \includegraphics[width=0.98\textwidth]{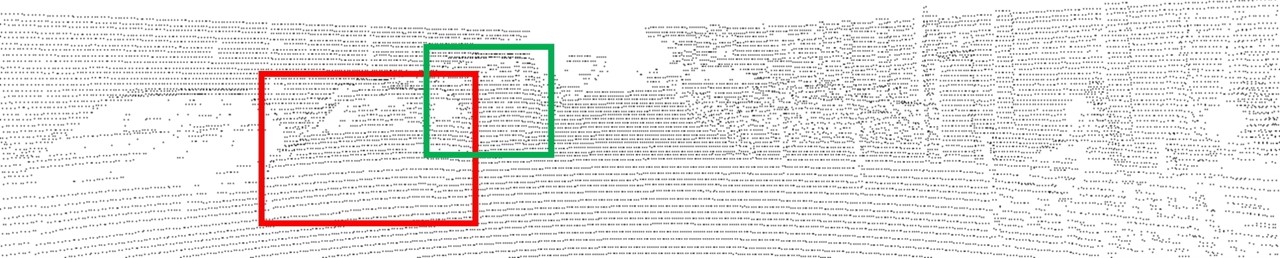}
        \caption{dense depth map scenario 3}
    \end{subfigure}%
    \newline
    \begin{subfigure}{0.32\textwidth}
        \includegraphics[width=0.98\linewidth]{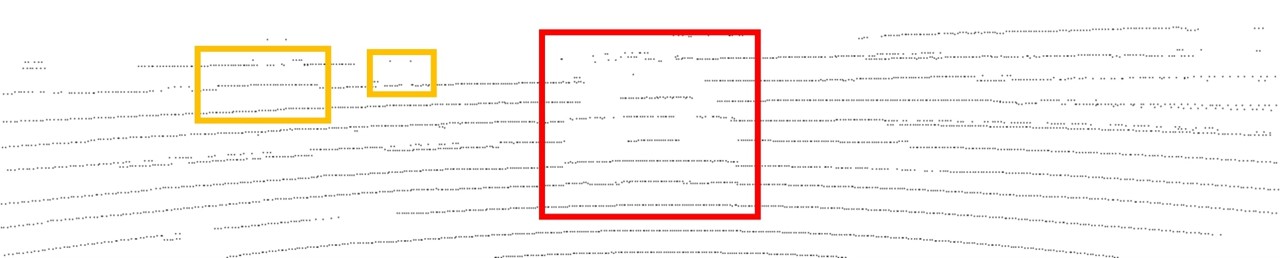}
        \caption{sparse depth map scenario 1}
    \end{subfigure}%
    \begin{subfigure}{0.32\textwidth}
        \includegraphics[width=0.98\linewidth]{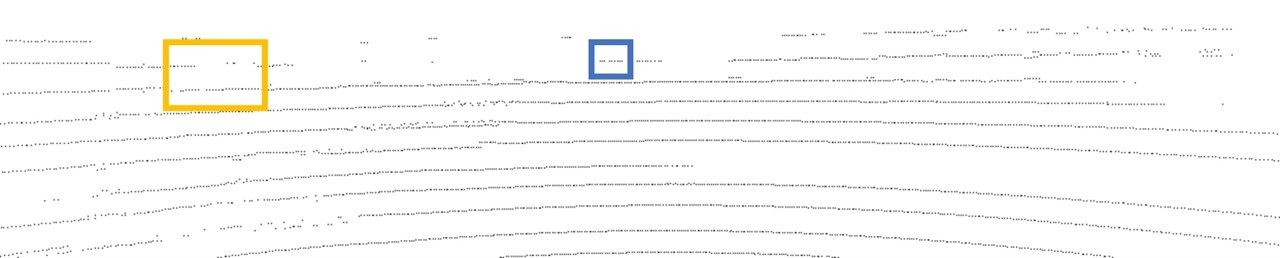}
        \caption{sparse depth map scenario 2}
    \end{subfigure}%
    \begin{subfigure}{0.32\textwidth}
        \includegraphics[width=0.98\linewidth]{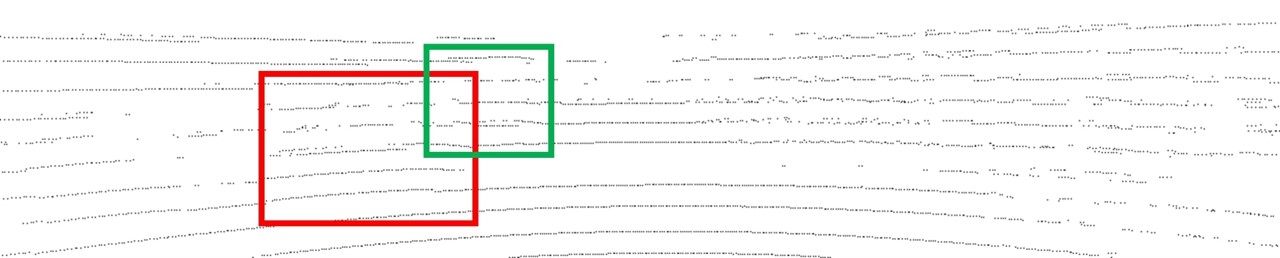}
        \caption{sparse depth map scenario 3}
    \end{subfigure}%
    \caption{Comparison of depth map from 16-line LiDAR (bottom) and 64-line LiDAR (middle) to their RGB image (top), on which  red boxes represent the short-range vehicles, orange boxes show the medium range vehicles and the far vehicles are marked by blue boxes. Green boxes illustrate the occluded vehicles.}
    \label{fig:dmap_comp}
\end{figure*}

In recent years, much research has been focused on autonomous driving technology. LiDAR is one of the most important sensors for perception tasks such as drivable region segmentation, object detection and vehicle tracking. Different from images captured by cameras, point cloud generated by LiDARs supplies 3D spatial information of the objects in the form of (X, Y, Z) coordinates and intensity. This alleviates the barrier of distance estimation and makes 3D object detection or tracking much more accurate. However, the price of high-resolution LiDARs is much higher than their low-resolution counterparts. The specifications of the most popular Velodyne 64-line LiDAR HDL-64E and 16-line LiDAR VLP-16 are compared in Table~\ref{tab:lidar_cmp}. As we can see, the cost of a low-resolution LiDAR is only about 1/18 of the high-resolution ones. Therefore, it is more economical to consider low-resolution LiDARs in order to build low-cost autonomous driving systems. However, it is a major challenge to perform object detection from the point cloud produced by a low-resolution LiDAR since it is too sparse to even show the shapes of objects. As illustrated in Fig.~\ref{fig:dmap_comp}, we can barely find objects from the depth map captured from a 16-line LiDAR, while in the 64-line LiDAR objects are more visible. 

\begin{table}[htbp]
    \centering
    \caption{A comparison of VLP-16 and HDL-64E LiDARs}
    \label{tab:lidar_cmp}
    \begin{tabular}{|l|l|l|} 
    \hline
    \textbf{LiDAR type} & \textbf{VLP-16} & \textbf{HDL-64E} \\
    \hline
    \hline
    Channel number & 16 & 64 \\
    \hline
    Res.(vertical / horizontal) & 2\textdegree / 0.2\textdegree & 0.4\textdegree / 0.1728\textdegree\\
    \hline
    Power (Watts) & 8 & 60 \\
    \hline
    Price (USD) & \$4'000 & \$75'000 \\
    \hline
    \end{tabular}
\end{table}

\section{RELATED WORKS}

\subsection{Low-Resolution LiDAR and Depth Completion} 
Some research works focused on segmentation using low-resolution LiDARs. \cite{gigli2020road} introduced the local normal vector for the LiDAR’s spherical coordinates as an input channel. Based on the existing LoDNN architectures \cite{calt2017fast}, its road segmentation performance using low-resolution LiDAR was close to that from high-resolution LiDAR within a reasonable degradation. A supervised domain adaptation was utilized by \cite{elhadidy2020improved} to predict the low-resolution point cloud into high-resolution point cloud in spherical coordinate and further evaluated the results in 3D semantic segmentation task. Low-resolution LiDARs had been also employed for object tracking tasks. In \cite{pino2017low}, a LiDAR-based system was proposed for estimation of actual positions and velocities of the detected vehicles.
Some other works utilized depth completion for 2D object detection, such as \cite{seikavandi2020deep} and \cite{farahnakian2020fusing}. In \cite{seikavandi2020deep}, a weighted depth filling algorithm was proposed to make the high-resolution (HDL-64E) LiDAR depth map even denser. Subsequently, this dense depth map was concatenated with the corresponding RGB image as the input of YOLOv3 \cite{redmon2018yolov3} network for 2D object detection. Similarly, the authors of \cite{farahnakian2020fusing} introduced a self-supervised depth completion network to fill the high-resolution depth map before detection 2D objects.

\subsection{High Resolution LiDAR for BEV Object Detection} Nearly all state-of-the-art object detectors utilize high-resolution LiDAR. In \cite{beltran2018birdnet}, it first transformed the point cloud into \textcolor{black}{Bird-Eye View (BEV)} map, and then extracted the ground and proposed the objects in two branches separately. Finally, the objects were predicted by a post-processing block. \cite{barrera2020birdnet+} further refined the previous version into an end-to-end model and achieved better performance. Single-stage detector, PIXOR, was proposed in \cite{yang2018pixor} by using 2D convolution on the voxelized BEV map. Without any anchor, it achieved real-time processing speed.

As mentioned earlier, due to the extreme sparsity, low-resolution LiDAR depth map does not supply enough shape information of the objects, but some sub-samples of the precise depth information. Meanwhile, the RGB image supplies rich context information. Thus, we argue that when fusing sparse depth map and RGB image together, 3D object detection becomes possible. 

\section{PROPOSED 3D OBJECT DETECTION FRAMEWORK}

\begin{figure}[htbp]
    \centering    
    \includegraphics[width=\linewidth]{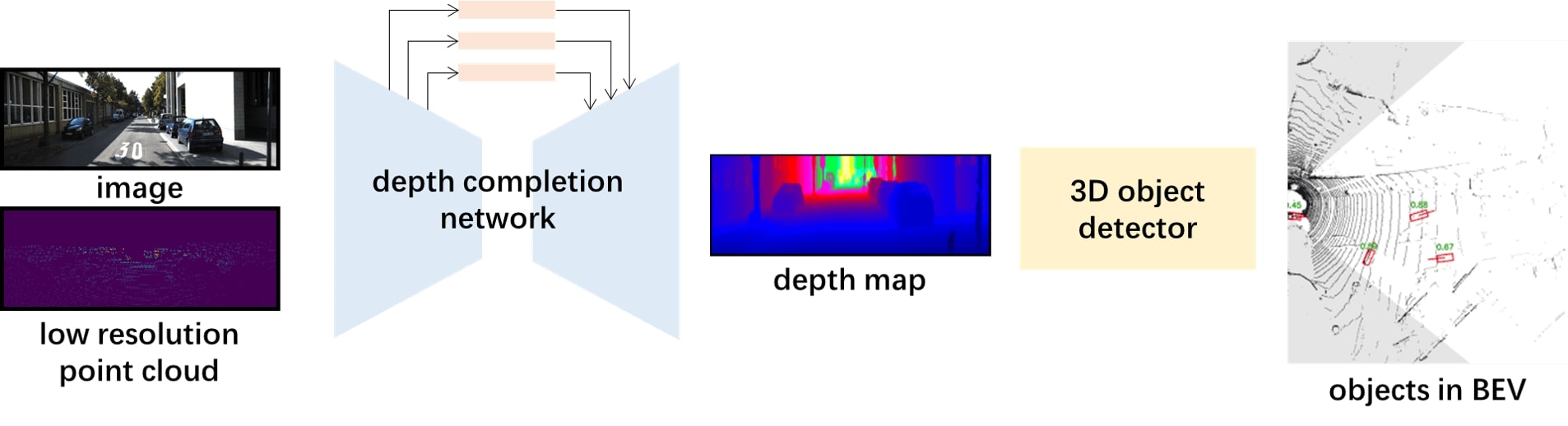}   
    \caption{The proposed framework for 3D object detection using low-resolution point cloud and RGB image}
    \label{fig:framework}     
\end{figure}

In this paper, we investigate the possibility of low-resolution LiDAR usage in BEV object detection task. In Fig.~\ref{fig:dmap_comp}, red box, orange box and blue box represent the vehicle in short range, medium range and long range respectively. For short range vehicles, their shapes are clearly visible from dense depth maps. In sparse depth maps, the shapes are very blurry but still recognizable since the number of points hitting on the vehicles is still large enough. Concerning the medium and long range vehicles (in orange and blue boxes), we can only get a small number of points even using 64-line LiDAR. While in the sparse depth map from 16-line LiDAR, the number of hit points is few to none. Taking the medium range vehicles in orange boxes in Fig.~\ref{fig:dmap_comp} (h) for example, it is easy to recognize them as obstacles due to sharp distance distinction but difficult to recognize them as vehicles. This also applies to vehicles with occlusion (green boxes in Fig.~\ref{fig:dmap_comp} (c), (f) and (i)). The long range vehicles in blue box (in both Fig.~\ref{fig:dmap_comp} (e) and (h)) get too few points to be correctly localized and classified. According to the analysis above, we found that unlike the depth map from 64-line LiDAR, 16-line LiDAR depth map does not show reliable context information but accurate distance information. This implies that 16-line LiDAR depth map is more useful for depth estimation rather than context information extraction. Therefore, to better use the information from 16-line depth map, we put a depth completion network prior to the object detector to generate a dense depth map with context information. After the dense depth map is generated, it is sent to 3D object detector, as demonstrated in Fig.~\ref{fig:framework}.


\subsection{Depth Completion Network}\label{sec:depth_com}
The depth completion network aims to fill the sparse depth map from 16-line LiDAR point cloud with the help of RGB image. The state-of-the-art depth completion network \cite{ma2019self} is adopted here with some modifications. It requires two inputs, RGB image and low-resolution sparse depth map. The RGB image supplies the context information in detail, while the sparse depth map supplies the precise depth information for some pixels on the image. The sensor fusion strategy adopted here is also referred as early fusion. To make the network more compact, we first replace the ResNet-34 backbone with ResNet-18. For performance improvement, global attention modules and an Atrous Spatial Pyramid Pooling (ASPP) \cite{chen2018deeplabv3+} module are placed to bridge the encoder and decoder. 

\begin{figure}[htbp]
    \centering    
    \includegraphics[width=0.6\linewidth]{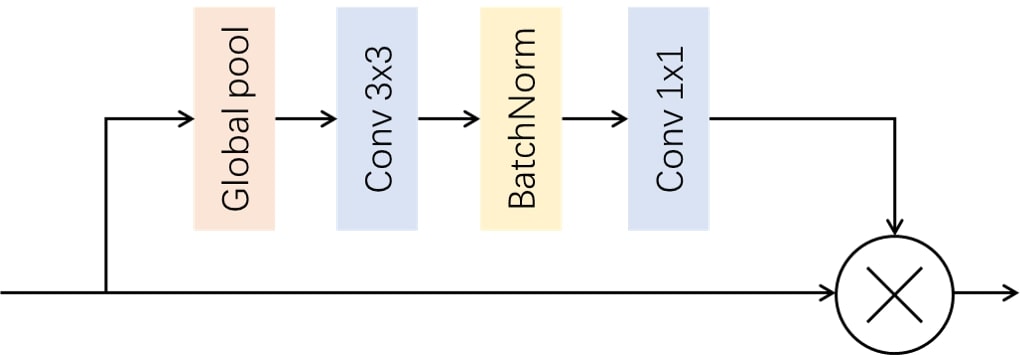}   
    \caption{Structure of global attention module}     
    \label{fig:gam}     
\end{figure}

As shown in Fig.~\ref{fig:gam}, the global attention module is used to extract global context information of the feature map by global pooling layer, and then fuse the global information back to guide the feature learning. Through adding this module, the global information is merged into features without up-sampling layer. This helps the decoder part to achieve better performance. Besides, an ASPP module is placed between encoder and decoder, with each convolution dilated rate 2, 4, 8 and 16. The ASPP module concatenates feature maps with different fields of perception, so that decoder has a better understanding of the context information.

The loss function of depth completion network is the Mean Square Error (MSE) between the predicted depth map and the ground truth.


\subsection{Object Detection Network}
The object detection network used in this framework is PIXOR \cite{yang2018pixor}. Its main idea is to take advantage of 2D convolution and anchor-free network to realize super-fast point cloud object detection in BEV. PIXOR consists of two steps. The first step is to reform the representation of input point cloud. It reduces 3 degrees of freedom to 2 in BEV, and extracts the 3rd freedom (z or height) as another input feature map channel. So that 2D convolution instead of 3D convolution is necessary to greatly decrease the computation complexity. The second step is to feed the reformed input feature map into an anchor-free one-stage object detector network. For the highly efficient computation on dense predictions, a fully convolutional architecture is utilized to build the backbone and header of PIXOR. Without any pre-defined anchors and proposals, PIXOR outputs the predicted class and orientation from header in a single network.


Concerning the loss function, the total loss of object detection consists of the classification loss and the regression loss (Eq.~\ref{eq:detect}), where $\lambda_{cls}$ and $\lambda_{reg}$ are the corresponding coefficients. The classification loss $\mathcal{L}_{cls}$ targets to correctly predict the object (cars in our case) and the regression loss $\mathcal{L}_{reg}$ aims to refine the size, center and the orientation of the predicted bounding boxes.

\begin{equation}
    \mathcal{L}_{detect} = \lambda_{cls}\mathcal{L}_{cls} + \lambda_{reg}\mathcal{L}_{reg}
    \label{eq:detect}
\end{equation}

\subsection{Implementation Details}
The depth completion network is first trained on KITTI Depth Completion dataset. The depth completion network is trained with batch size of 4, and learning rate starts at 1e-4 which decreases every 5 epochs. The total number of training epoch is 10. After training the depth completion network and keeping as it is, we move on to train the object detector from scratch. The KITTI Object Detection dataset has been split into training and validation parts according to \cite{qian2020end}.  The optimizer is Adam, with batch size 8. The learning rate starts at 1e-3 and reduces by a factor of 2 when the validation loss does not decrease. Finally, we fine-tune the entire framework with both depth completion network and object detection network together, with 16-line point cloud and images as input and vehicles in BEV as output. 

\section{EXPERIMENT RESULTS}

\subsection{Evaluation Dataset}
Training and evaluation of the whole framework both employ KITTI dataset (both Depth Completion and Object Detection). Before feeding into the framework mentioned above, the point clouds are down-sampled to emulate the VLP-16 low-resolution LiDAR. KITTI depth completion dataset contains 85,898 training data and 1,000 selected validation data. Its ground truth is produced by aggregating consecutive LiDAR scan frames into a semi-dense depth map, about 30\% annotated pixels. KITTI object detection dataset has 7,481 training data and 7,518 testing data. Evaluation is categorized into three regimes: easy, moderate and hard, representing objects at different occlusion and truncation levels.

\subsection{Depth Completion Performance Evaluation}
As described in Sec.~\ref{sec:depth_com}, in order to enhance the depth completion performance, multiple GAM modules have been added to bridge the encoder and the decoder of depth completion network. The performance comparison on validation dataset is illustrated in Tab.~\ref{tab:dc_perform}. Adding GAM modules results in the performance improvement of about 3.6\% and 7.0\% measured by Root Mean Square Root (RMSE) and Mean Average Error (MAE) respectively.

\begin{table}[htbp]
    \centering
    \caption{Depth completion performance comparison with and without GAM modules}
    \label{tab:dc_perform}
    \begin{tabular}{|c|c|c|}
    \hline
    with GAMs & RMSE (1/mm) & MAE (1/mm) \\
    \hline
    Yes & 1592.74 & 537.81\\
    \hline
    No & 1651.68 & 578.14\\
    \hline
    \end{tabular}
\end{table}

\begin{figure}[htbp]
    \centering
    \begin{subfigure}{0.5\linewidth}
        \includegraphics[width=0.98\linewidth]{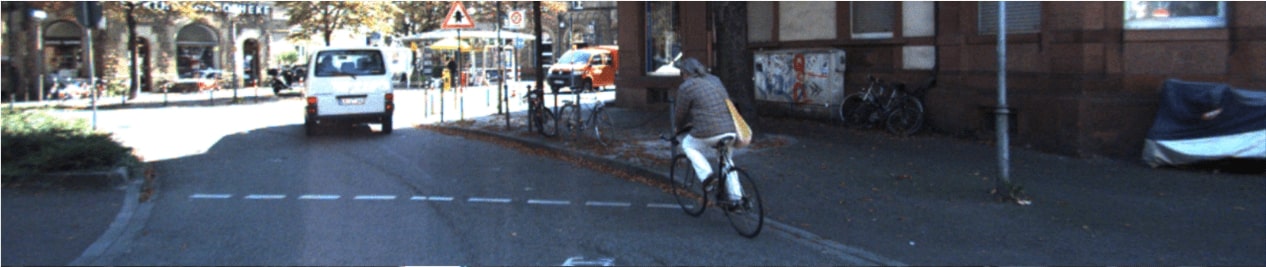}
        \vspace{-0.1cm}
        \caption{Image}
        \includegraphics[width=0.98\linewidth]{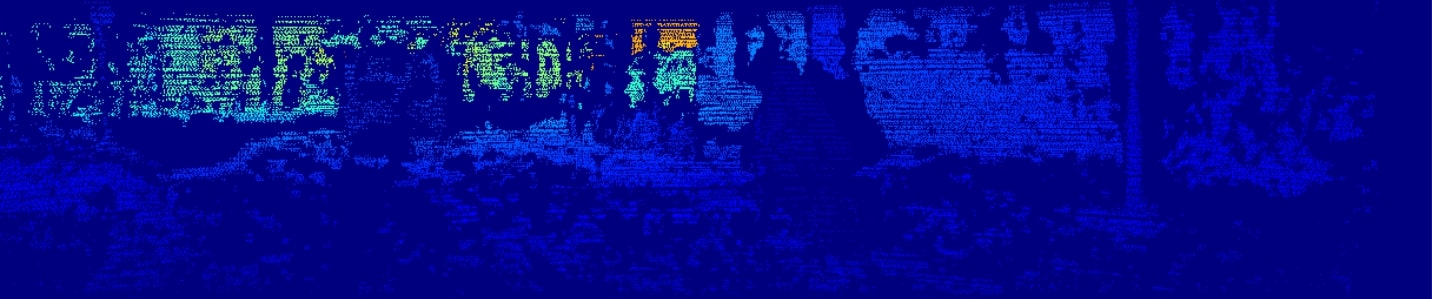}
        \vspace{-0.1cm}
        \caption{Depth ground truth}
        \captionsetup{labelformat=empty}
    \end{subfigure}%
    \begin{subfigure}{0.5\linewidth}
        \includegraphics[width=0.98\linewidth]{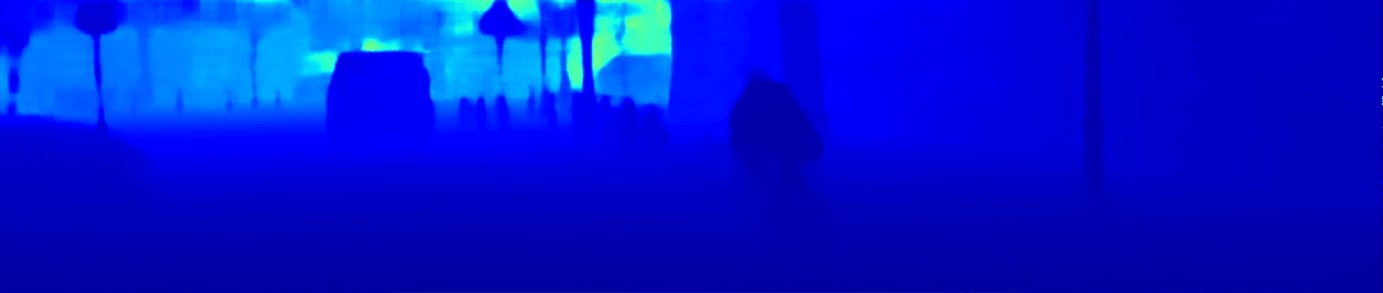}
        \vspace{-0.1cm}
        \caption{Depth map without GAM}
        \includegraphics[width=0.98\linewidth]{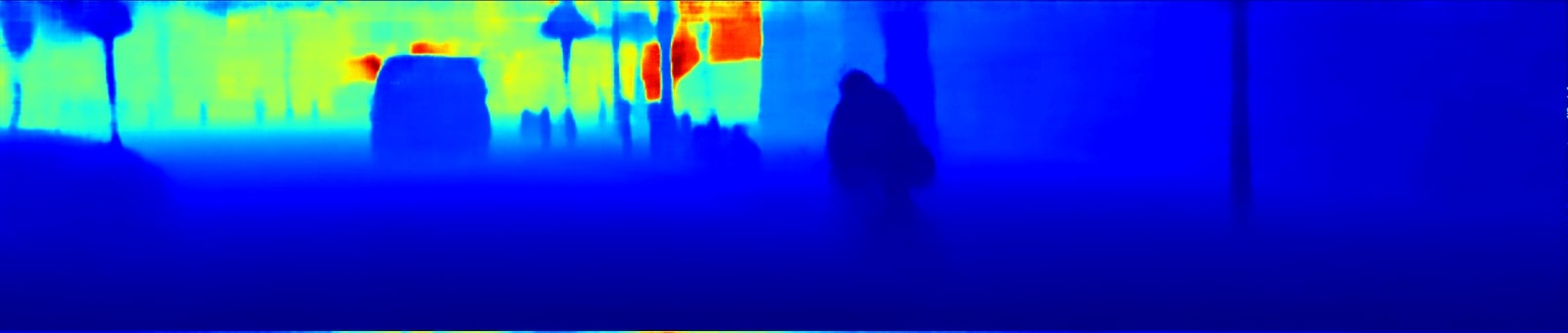}
        \vspace{-0.1cm}
        \caption{Depth map with GAM}
    \end{subfigure}%
    \caption{Comparison of depth map from 16-line LiDAR with and without GAM module to their RGB image and ground truth}
    \label{fig:dc_cmp}
    \vspace{-0.6cm}
\end{figure}

Fig.~\ref{fig:dc_cmp} (b) and (c) demonstrate the predicted depth maps of depth completion networks with and without GAM modules respectively. And the bottom figure shows the ground truth. In this example, the depth map from depth completion network with GAM module gives objects slightly better shape representation.

%
%
%

\begin{table*}[htbp]
    \centering
    \caption{\textbf{BEV performance comparison on KITTI object detection validation dataset.} This table shows $AP_{BEV}$ (in \%) of the car category, corresponding to average precision of the bird's-eye view.}
    \begin{tabular}{|c|c|c|c|c|c|c|c|}
    \hline
    \multirow{2}*{Detection Networks} & \multirow{2}*{Input} & \multicolumn{3}{|c|}{IoU=0.5} & \multicolumn{3}{|c|}{IoU=0.7} \\
    \cline{3-8}
     & & Easy & Moderate & Hard & Easy & Moderate & Hard \\
    \hline
    PIXOR\cite{yang2018pixor} & LiDAR only (64-line) & 94.2 & 86.7 & 86.1 & 85.2 & 81.2 & 76.1 \\
    \hline
    PIXOR & LiDAR only (16-line) & 60.7 & 51.2 & 46.8 & 53.8 & 47.1 & 39.1 \\
    \hline
    \textbf{Ours} & \textbf{LiDAR (16-line) + Camera} & \textbf{89.0} & \textbf{75.8} & \textbf{68.1} & \textbf{75.4} & \textbf{61.2} & \textbf{55.2} \\
    \hline
    \end{tabular}
    \label{tab:results}
\end{table*}

%
%
%

\begin{figure}[htbp]
    \centering
    \begin{subfigure}{0.5\linewidth}
        \includegraphics[width=0.98\linewidth]{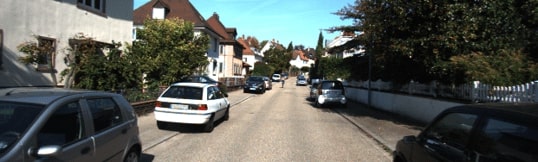}
        \includegraphics[width=0.98\linewidth]{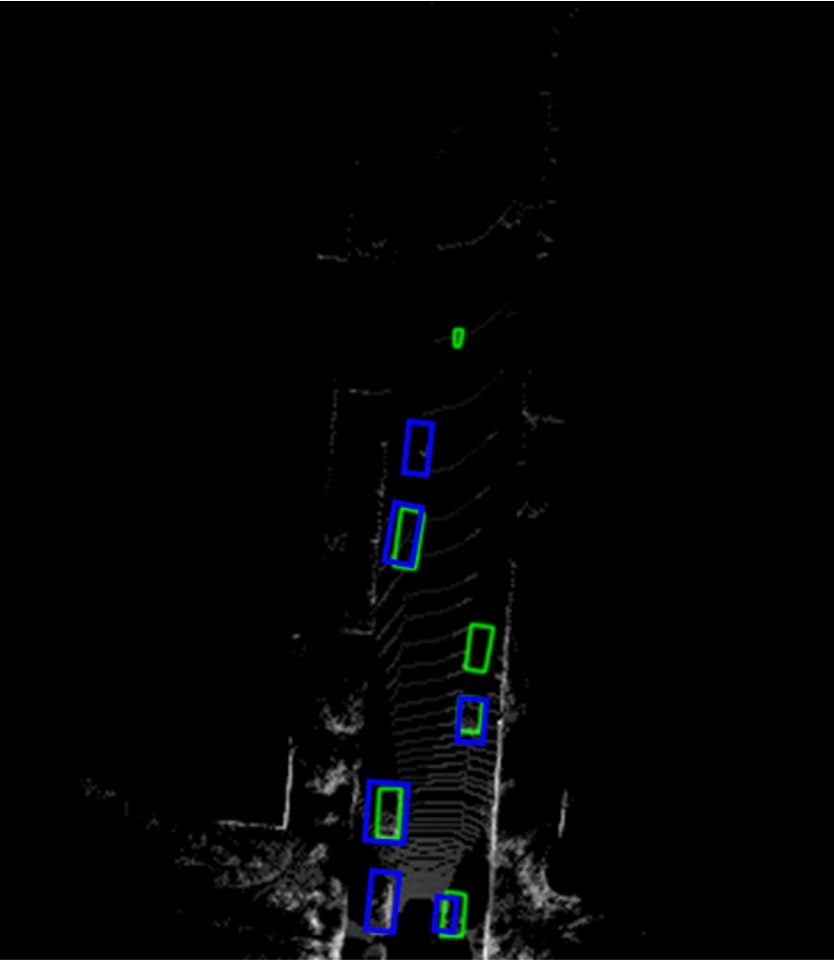}
    \end{subfigure}%
    \begin{subfigure}{0.5\linewidth}
        \includegraphics[width=0.98\linewidth]{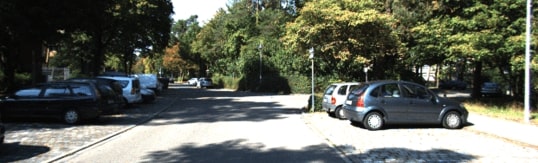}
        \includegraphics[width=0.98\linewidth]{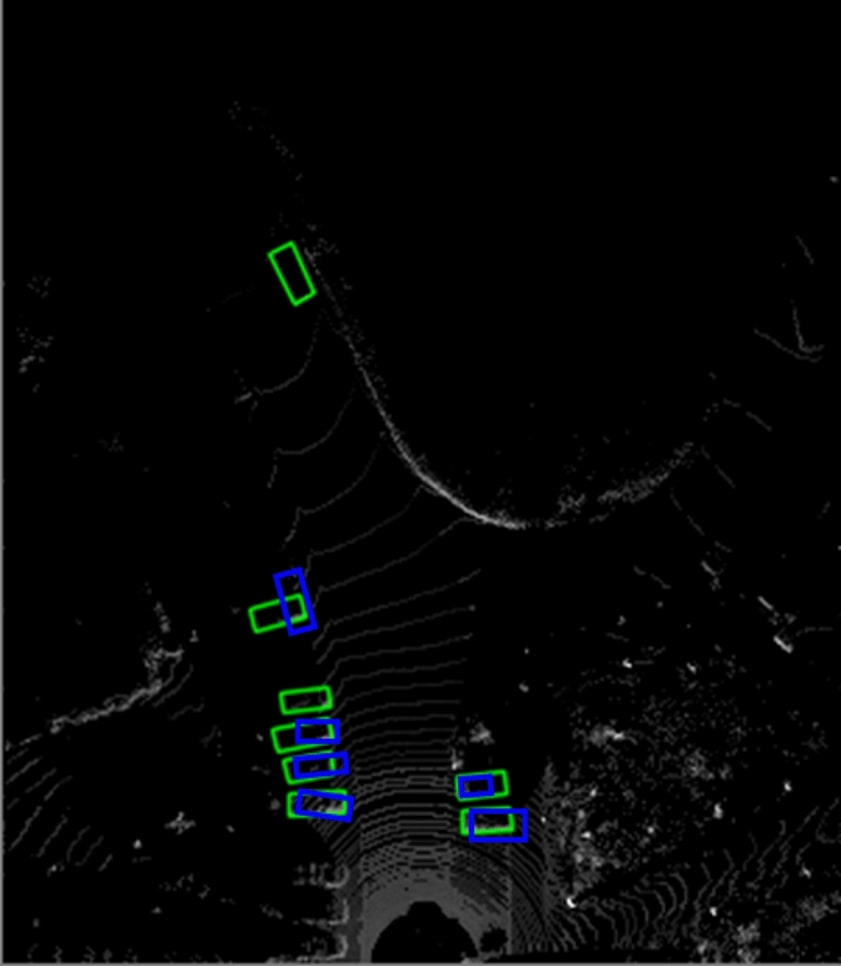}
    \end{subfigure}%
    \newline
    \begin{subfigure}{0.5\linewidth}
        \includegraphics[width=0.98\linewidth]{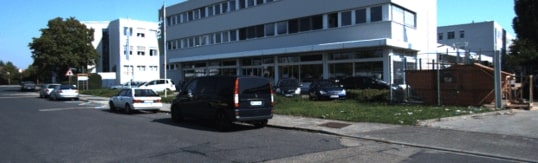}
        \includegraphics[width=0.98\linewidth]{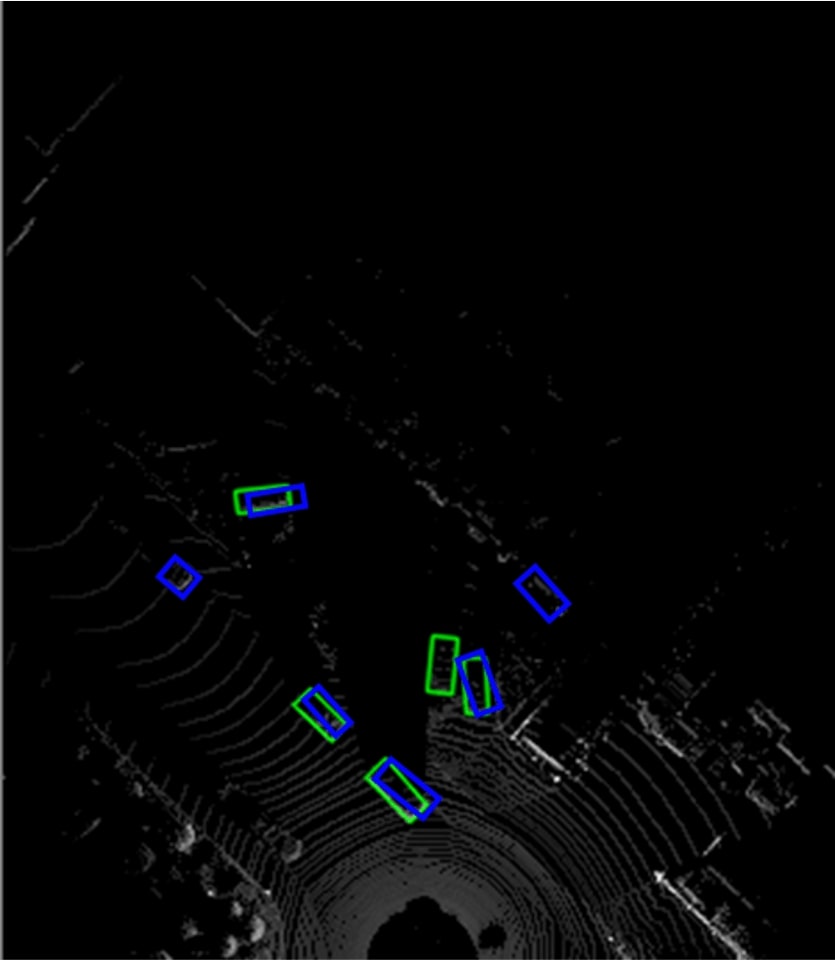}
    \end{subfigure}%
    \begin{subfigure}{0.5\linewidth}
        \includegraphics[width=0.98\linewidth]{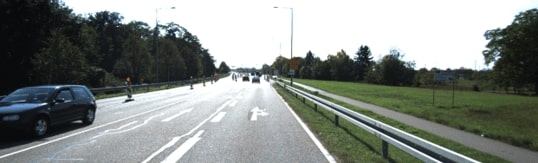}
        \includegraphics[width=0.98\linewidth]{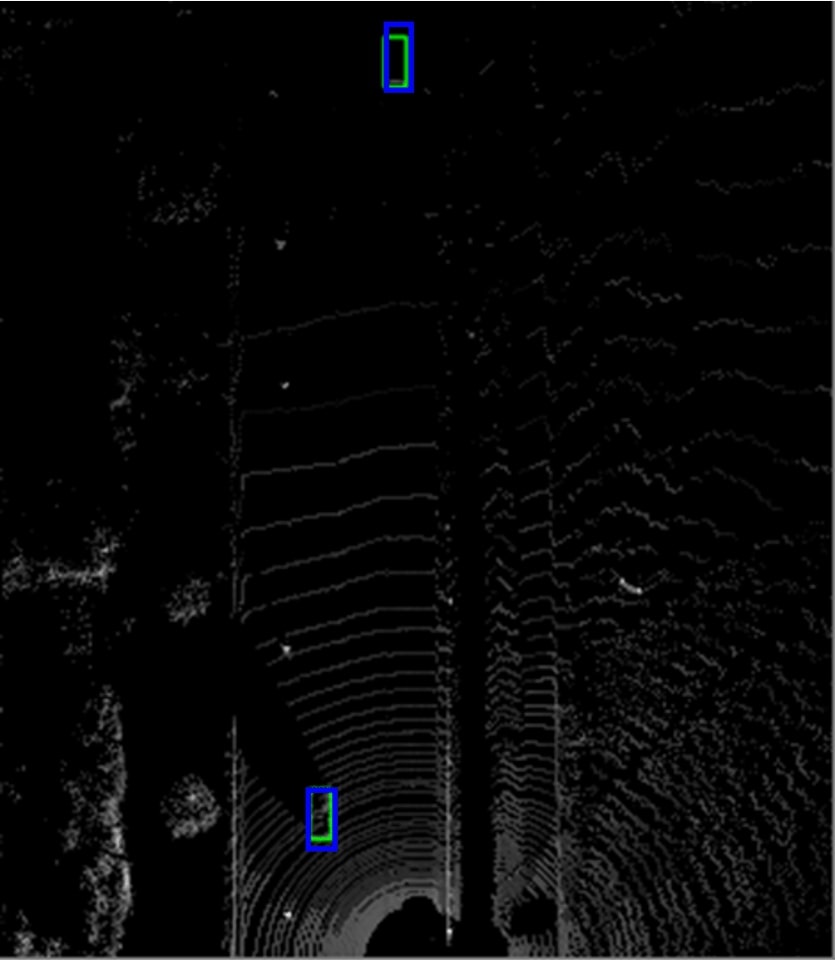}
    \end{subfigure}%
    \caption{Visualization of object detection from the proposed framework, where the green boxes are ground truth and the blue boxes represent the predicted results}     
    \label{fig:result}     
\end{figure}

\subsection{Object Detection Performance Evaluation}
\begin{figure}[htbp]
    \centering
    \includegraphics[width=0.9\linewidth]{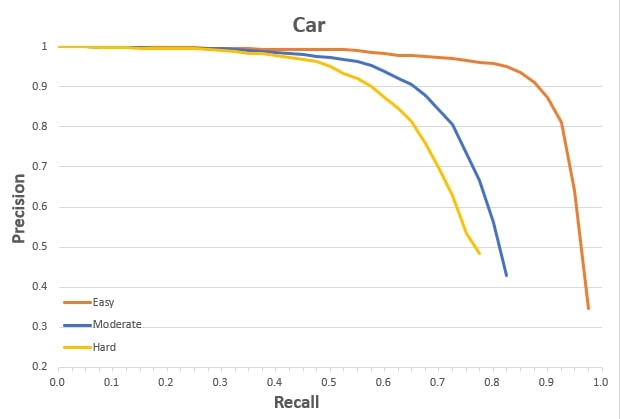}
    \caption{Precision-Recall curve of the proposed framework on KITTI val dataset.}
    \label{fig:curve}
\end{figure}

The performance of our framework on KITTI object detection validation dataset is illustrated in Tab.~\ref{tab:results} and Fig.~\ref{fig:result}. The results are shown in two circumstances IoU=0.5 and IoU=0.7 respectively. When IoU=0.5, our framework achieves 89.0\%, 75.8\% and 68.1\% detection accuracy for easy, moderate and hard cases respectively. While in case of IoU=0.7, the prediction accuracy is decreased to 75.4\%, 61.2\% and 55.2\% respectively. Compared to feeding 16-line point cloud directly into PIXOR, our framework pulls up the detection accuracy significantly in all cases. If compared to PIXOR with 64-line point cloud as input, the performance of our framework is relatively comparable in easy and moderate cases. But in hard case, the prediction accuracy drops around 20\% in both IoU criteria. The precision-recall curve is demonstrated in Fig.~\ref{fig:curve}.


\textcolor{black}{In regard to the computations, when running on RTX 2080Ti, the inference time of proposed network is 25.4ms or 39.8 frames per second (fps). Besides, as the network is aiming for embedded systems, we also tested it on DRIVE PX2 that contains two discrete Pascal GPUs. The inference latency for each point cloud frame is 581.7ms. If two GPUs run as two threads, the throughput increases to 3.4fps.}

\section{Conclusion}
This paper presents a framework that enables 3D object detection using a low-resolution LiDAR. By cascading a depth completion network with an object detector, it first converts the sparse point cloud into a denser depth map that is subsequently processed for 3D object detection. It demonstrates 3D object detection with only a 16-line LiDAR and a camera. When evaluated on KITTI dataset, the proposed solution achieves high accuracy in object detection for both easy and moderate cases, comparable to the benchmarks using 64-line LiDARs.

\end{document}